# Integrating Case Based And Rule Based Reasoning: The Possibilistic Connection

Soumitra Dutta
INSEAD, Fontainebleau
France 77305
Dutta%Freiba51.bitnet

Piero P. Bonissone
Artificial Intelligence Program
General Electric Corporate R&D
Schenectady, New York 12301
Bonissone@crd.ge.com

## Abstract

Rule based reasoning (RBR) and case based reasoning (CBR) have emerged as two important and complementary reasoning methodologies in artificial intelligence (AI). For problem solving in complex, real world situations, it is useful to integrate RBR and CBR. This paper presents an approach to achieve a compact and seamless integration of RBR and CBR within the base architecture of rules. The paper focuses on the possibilistic nature of the approximate reasoning methodology common to both CBR and RBR. In CBR, the concept of similarity is casted as the complement of the distance between cases. In RBR the transitivity of similarity is the basis for the approximate deductions based on the generalized *modus ponens*. It is shown that the integration of CBR and RBR is possible without altering the inference engine of RBR.

This integration is illustrated in the financial domain of mergers and acquisitions. These ideas have been implemented in a prototype system, called MARS.

## 1 Introduction

In this section, we introduce rule based and case based reasoning methodologies, describe their treatment of uncertain information, emphasize the need for their integration and outline the focus and structure of the paper.

### 1.1 Rule Based and Case Based Reasoning

Rule based reasoning (RBR) [BS84] is one of the most popular reasoning paradigms used in artificial intelligence (AI). The reasoning architecture of rule based systems has two major components: the *knowledge base* (usually consisting of a set of "IF...THEN..." rules representing domain knowledge) and the *inference engine* (usually containing some domain independent inference mechanisms, such as forward/backward chaining).

Case based reasoning (or analogical reasoning), though common and extremely important in human cognition, has only recently emerged as a major reasoning methodology. Case based reasoning (CBR) involves solving new problems by identifying and adapting *similar* problems stored in a library of past experiences/problems. The reasoning architecture of CBR consists of a *case library* (stored representations of previous experiences/problems solved) and an *inference cycle*. The important steps in the inference cycle of CBR are to *find* and *retrieve* cases from the case library which are most relevant to the problem at hand (input) and to *adapt* the retrieved cases to the current input. Within this broad framework, two major classes of CBR can be identified [RS89b]: *problem solving CBR* and *precedent based CBR*. In problem solving CBR, the emphasis is on adapting the retrieved cases to find a plan or a course of action to solve the input problem (such as in industrial design and planning [BM88]). In contrast, the focus in precedent based CBR is to use the retrieved cases to justify/explain an action/solution (common in legal reasoning [AR88]).

### 1.2 Uncertainty in RBR and CBR

Uncertainty is pervasive in the reasoning cycle of deductive (rule-based) and analogical (case-based) reasoning systems.

In Rule-Based Reasoning (RBR), uncertainty is present in the knowledge used in the task. Rules elicited from domain experts are usually *plausible* rather than *categorical* in nature. The partial degree of belief entailed by these rules is propagated through the inference network to determine the degree of confirmation and refutation of the various conclusions.

In CBR, uncertainty is present in the semantics of *abstract features* used to index the cases, in the evaluation and (hierarchical) aggregation of the *similarity measures* computed across these features, in the determination of *relevancy and saliency* of the similar cases, and in the solution adaptation phase.

In Section 2.2 we will show how most of this uncertainty can be modeled by using fuzzy predicates and plausible rules to derive abstract features from the surface features. Similarity measures can be defined as the complement of metrics between fuzzy-sets (cases). The similarity measure can be aggregated or chained (using the transitivity of similarity) according to well-defined operators (Triangular norms.)

### 1.3 Integration of Reasoning Methodologies

The need to integrate diverse reasoning techniques for effectively solving complex real world problems has been recently recognized by the AI community. This is represented in the works of Carbonell and Velose [CV88] (integration of CBR and classical search problem solvers), Hammond and Hurwitz [HH88] (integrating CBR and explanation based reasoning) and Rissland and Skalak [RS89b] (integrating RBR and CBR).

RBR and CBR are largely *complementary* reasoning methodologies. RBR can better represent specialized domain knowledge in a modular, declarative fashion, while CBR can better represent past experiences and domain complexity [RS89a]. Significant benefits are possible by combining RBR and CBR. For example, CBR can directly enhance RBR by providing a context for screening the knowledge base and by extending the coverage of rules by representing exceptions (to the rule) in the form of cases. Going the other direction, RBR can enhance CBR by expressing domain knowledge to dynamically determine the contextually dependent relevance of a feature set (or attributes of a case) to a given goal and to dynamically select the best similarity/relevancy measure to use for case retrieval. There are numerous domains in which it is important to combine RBR and CBR, e.g., the legal domain (see Section 3.2 for an example).

### 1.4 Focus and Structure of Paper

This paper is concerned with the integration of RBR and CBR in an uncertain and dynamic world. Rather than "patching together" two different types of representational and reasoning frameworks, we have chosen to attempt the integration within one architectural framework, namely that of RBR. As shown in later sections, it is possible to achieve a compact, seamless integration of the two reasoning methodologies without changing the inference engine of RBR. This has, as discussed later, some advantages over other architectures (e.g., the one of Rissland and Skalak [RS89b]. The incorporation of uncertainty into the reasoning framework gives the system added power in handling real world situations, which are almost invariably uncertain and dynamic. It also leads to a more accurate treatment of CBR, as it is inherently a non-deductive form of approximate reasoning in which there is significant uncertainty and imprecision, e.g., in the semantics of the case features and in determining the similarity/relevancy of prior cases to the input problem/goal. The significance of our work arises from the fact that though RBR and CBR are two extremely important reasoning methodologies, there has been very little research in combining the two.

The domain chosen for the illustration of our ideas is the financial domain of mergers and acquisitions (M&A). M&A represent a real world situation, which is complex, uncertain, dynamic and relevant for business today. The ideas and technical approach detailed in this paper have been implemented in a prototype system, called MARS [BD90].

This paper contains four other sections. Section 2 illustrates the role of approximate reasoning techniques and contrasts probabilistic and possibilistic reasoning systems. A brief introduction to the domain of M&A is provided in Section 3. The need for integrating RBR and CBR in this domain is also explained in that section. Section 4 provides an overview of MARS, illustrates the nature of possibilistic reasoning in MARS, and then describes the integration of RBR and CBR in MARS. Section 5 compares our work with related research and concludes the paper by describing the contributions and limitations of this research.

## 2 Approximate Reasoning Systems

The task of a reasoning system is to determine the *truth value* of statements describing the state or the behavior of a real world system. However, this hypothesis evaluation requires complete and certain information, which is typically not available. Therefore, approximate reasoning techniques are used to determine a *set of possible worlds* that are logically consistent with the available information. These possible worlds are characterized by a set of propositional variables and their associated values. As it is generally impractical to describe these possible worlds to an acceptable level of detail, approximate reasoning techniques seek to determine some properties of the set of possible solutions or some constraints on the values of such properties [Rus90b].[1]

A large number of approximate reasoning techniques have been developed over the past decade to provide these solutions. (See references [Bon87a], [Pea88] for a survey). These techniques can be roughly subdivided into two basic categories ac-cording to their ***quantitative*** or ***qualitative*** characterizations of uncertainty. Among the quantitative approaches, we find two types of reasoning that differ in the semantics of their numerical representation. One is the ***probabilistic reasoning*** approach, based on probability theory. The other one is the ***possibilistic reasoning*** approach, based on the semantics of many-valued logics. We will briefly contrast these two types of quantitative representations and focus our discussion on possibilistic reasoning systems.

[1] The authors want to acknowledge Enrique Ruspini's private communication, which is the basis for the content of this section. The interested reader should consult reference [Rus87], [Rus89], [Rus90b] for a further elaboration of this topic.

## 2.1 Probabilistic and Possibilistic Reasoning Systems

Probability-based reasoning, or ***probabilistic reasoning*** seeks to describe the constraints on the variables that characterize the possible worlds with conditional probability distributions based on the evidence in hand. Their supporting formalisms are based on the concept of ***set-measures***, additive real functions defined over certain subsets of some space.

These methods focus on chance of occurrence and relative likelihood. They are oriented primarily toward the choice of decisions that are optimal in the ***long-run***, as they measure the ***tendency*** or ***propensity*** of truth of a proposition without assuring its actual validity. Thus, probabilistic reasoning estimates the frequency of the truth of a hypothesis as determined by prior observation (objectivist interpretation) or a degree of gamble based on the actual truth of the hypothesis (subjectivist interpretation).

Possibilistic reasoning, which is rooted in fuzzy set theory [Zad65] and many-valued logics, seeks to describe the constraints on the possible worlds in terms of their ***similarity*** to other sets of possible worlds.

These methods focus on ***single*** situations and cases. Rather than measuring the tendency of the given proposition to be valid, they seek to find another proposition that resembles (according to some measure of similarity) the hypothesis of interest but that is valid. Thus, possibilistic reasoning asserts that a related, similar (and usually less specific) hypothesis is true.

## 2.2 Possibilistic Reasoning

Given the purpose and characteristics of probabilistic and possibilistic reasoning, it is clear that these technologies ought to be regarded as being complementary rather than competitive.

The single-case orientation of possibilistic techniques makes them particularly suitable for case-based reasoning. In CBR, it is typically the case that the problem in hand (probe) has never been encountered before. The inference in CBR is based on the existence of cases ***similar enough*** (i.e. close enough) to the probe to justify the adaptability of their solution to the current problem.

The notion of similarity is based on the concept of ***metric*** or distance, as opposed to that of set measure. Distances are functions which assign a number greater that zero to pairs of elements of some set (for sake of simplicity, we will assume the range of this function to be the interval [0,1]). Distances are ***reflexive, commutative***, and ***transitive***. Similarity can be defined as the complement of distance, i.e.:

$$S(A,B) = 1 - d(A,B)$$

The basic structural characteristics of the similarity functions is an extended notion of transitivity that allows the computation of bounds on the similarity between two objects A and B on the basis of knowledge of their similarities to a third object C:

$$S(A,B) \geq T(S(A,C), S(B,C)),$$

where T is a Triangular-norm [BD86], [Bon87b]. Any continuous triangular norm $T(A,B)$ falls in the interval $Max(0, A+B-1) \leq T(A,B) \leq Min(A,B)$. Thus, we can observe that the if we use the lower bound of the range of T-norms in the expression describing the transitivity of similarity, we obtain the triangular inequality for distances. If we use the upper bound, we obtain the ultrametric inequality.

This similarity notion is a direct extension of the notion of ***accessibility relation*** that is of fundamental importance in modal logics. This notion is further described by Ruspini in these proceedings [Rus90a]. In summarizing Ruspini's results, we can observe that the notion of accessibility captures the idea that whatever is true in some world $w$, is true, but in a modified sense, in another $w'$ that is accessible from it. When considering multiple levels of accessibility (indexed by a number between 0 and 1), this relation, measuring the resemblance between two worlds, may be used to express the extent by which considerations applicable in one world may be extended to another world.

The basic inferential mechanism, underlying the ***generalized modus-ponens*** [Zad79], makes use of inferential chains and the properties of a similarity function to relate the state of affairs in the two worlds that are at the extremes of an inferential chain.

We have briefly summarized the semantics of possibilistic reasoning, its role in determining the similarity between possible worlds (cases), and its mechanism to propagate similarities through a reasoning chain (rule chain). On this basis, we have established a common ground upon which we can build the integration of CBR and RBR. Before proceeding to describe such an integration, we need to justify the reasons for integrating these two methodologies. This motivation will be provided by the description of the problem domain of Mergers and Acquisition (M&A), which is used to test the integration.

## 3 Mergers & Acquisitions (M&A)

This section introduces the domain of M&A and emphasizes the need for integrating RBR and CBR in M&A.

### 3.1 Introduction and Overview

The structure of corporate USA has been changed dramatically by the flood of mergers and acquisitions witnessed over the past years. Today, a flurry of mergers are sweeping through European industry as it prepares for 1992. Annually, these deals total more than a few billions of US dollars. The average M&A deal is enormously complex and involves sophisticated reasoning and planning on the part of several parties. To lend some useful conceptual abstraction, we can consider two players of interest in simple M&A deals: the *raider* (who usually initiates a take-over attempt) and *the target* (which is the company of interest to the raider). Another player of interest who is outside the structure of the actual M&A deal, but has a keen interest in the entire process is the *professional arbitrageur*. While the actions of each of these players vary from deal to deal, it is possible to identify certain basic actions associated with their individual roles. For example, some of the representative actions of a raider are: *target monitoring, evaluation and selection; attack strategy selection; target's response evaluation and attack strategy adaptation*. Even in simple M&A deals, other complicating factors, such as multiple bidders and legal complications, often arise. The reader may consult the references [Fer87], [Mic86], [Roc87] for more details on various aspects of M&A.

### 3.2 The Anti-trust Defense

Usually, when a raider launches a hostile takeover attempt, the target has to devise an elaborate defense strategy. Michel and Shaked [Mic86] note that "*anti-trust arguments are one of the most frequently used forms of merger defense*". The laws governing anti- trust cases depend on several merger guide-lines (e.g., the 1982 and 1984 guide-lines) issued by the Department of Justice (DOJ) (in the USA). Much of the reasoning involved in the interpretation and application of guide-lines regarding anti-trust laws can be expressed by rules. For example, the 1982 guide-lines specified that markets where the post merger HHI (a mathematical measure of market concentration) was above 1800 were *highly concentrated* and if the post merger HHI was between 1000 and 1800, then the market was *moderately concentrated* and so on. However, such rules by themselves are not enough as [Mic86] "*it is not possible to remove the exercise of judgement from the evaluation of mergers under the anti-trust laws*". This exercise of judgement is predominant in resolving issues like definition and measurement of market, efficiency arguments and treatment of foreign competition. This is where CBR can help and is used extensively.

For example, consider the $5.1 billion attempt by Mobil to takeover Marathon on Nov. 1, 1981. Marathon began the takeover defense by filing an anti-trust lawsuit against Mobil (if successful, Mobil would then become the largest marketer of gasoline in the USA with an estimated 10% market share). The key issue here was whether section 7 of the Clayton Act (which provides that "*no person .. shall acquire .. stock .. where, in any line of commerce .. in any section of the country, the effect of such acquisition may be substantially to lessen competition*") was being violated by the merger. Judge J. M. Manos of the Ohio Court ruled against Mobil and in his ruling [Mic86] "*referred to past cases similar to the Mobil-Marathon situation. In the 1962 case of Brown Shoe Company, the combined market share of Brown Shoe and G.R. Kinney Co. was found to exceed 5% nationwide, rising to 57% in some cities. In the 1966 case of Pabst Brewing Company, the merged firm would have had a combined market share of 4.49% in the USA and up to 23.95% in Wisconsin.*" In all these three cases, section 7 of the Clayton Act was found to be violated.

This brief example illustrates the important role that cases and rules have in the M&A domain and stresses the need for their integration.

## 4 MARS: A Mergers & Acquisitions Reasoning System

In this section, we first provide a quick overview of MARS and then focus on the integration of RBR and CBR within MARS.

### 4.1 Overview of MARS

MARS [BD90] is a prototype AI reasoning system that both simulates and provides expert advice regarding the actions of the raider, the target and the arbitrageur. The general software architecture of MARS is as shown in Figure 1. There are four independent simulators. The global simulator provides a simulation of the variations of the macroeconomic variables affecting the M&A deal (e.g., the interest rate and the T-Bill price). The three other simulators simulate the reasoning and planning strategies of the raider, target and the arbitrageur respectively. There is a fusion of different reasoning techniques in all four simulators and each of them is independently capable of integrated reasoning and planning with uncertain, incomplete and time varying information.

MARS is implemented in Common LISP using KEE and Reasoning with Uncertainty Module

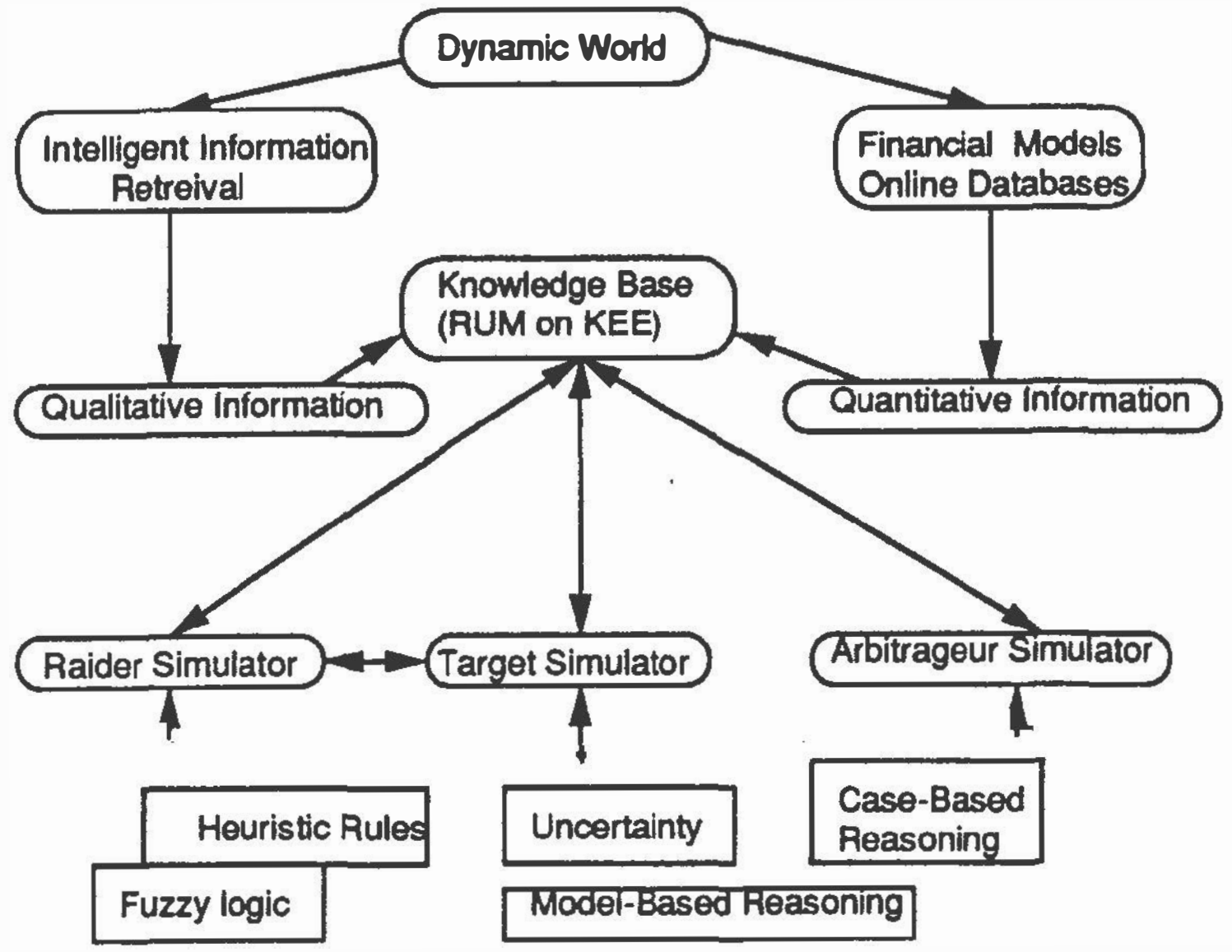

Figure 1: MARS Software Architecture

(RUM) [BGD87], and runs on the Symbolics. The knowledge base of MARS is frame based and consists of approximately 550 KEE units. Figure 2 shows the user interface for MARS. More details on the structure, implementation and use of MARS can be found in [BD90].

### 4.2 Possibilistic Reasoning in MARS

The *generalized modus ponens* and its associated possibilistic approach has been implemented in RUM, a reasoning shell described in [BGD87] and [BW89]. For the reader's convenience it is briefly summarized in this section.

Uncertainty in RUM is represented in both facts and rules. Facts are qualified by a degree of confirmation and a degree of refutation. For a fact A, the lower bound of the confirmation and the lower bound of the refutation are denoted by $L(A)$ and $L(\neg A)$ respectively. As in the case of Dempster's [Dem67] lower and upper probability bounds, the following identity holds: $L(\neg A) = 1 - U(A)$, where $U(A)$ denotes the upper bound of the uncertainty in $A$ and is interpreted as the amount of failure to refute $A$. Note that $L(A) + L(\neg A)$, need not necessarily be equal to 1, as there may be some ignorance about $A$ which is given by $(1 - L(A) - L(\neg A))$. The degree of confirmation and refutation for the proposition $A$ can be written as the interval $[L(A), U(A)]$.

RUM provides a natural representation for plausible rules. Rules are discounted by *sufficiency* ($s$), indicating the strength with which the antecedent implies the consequent and *necessity* ($n$), indicating the degree to which a failed antecedent implies a negated consequent. Note that conventional strict implication rules are special cases of plausible rules with S = 1 and N = 0. RUM's inference layer is built on a set of five Triangular norms (T-norms) based calculi [Bon87b]. T-norms and T-conorms are two-place functions from [0,1]X[0,1] to [0,1] that are monotonic, commutative and associative. They are the most general families of binary functions which satisfy the requirements of the conjunction and disjunction operators respectively. Their corresponding boundary conditions satisfy the truth tables of the logical AND and OR operators. Five uncertainty calculi based on the following five T-norms are used in RUM:

$$T_1(a,b) = max(0, a+b-1)$$

$$T_{1.5}(a,b) = (a^{0.5}+b^{0.5}-1)^2 \quad \text{if } (a^{0.5}+b^{0.5}) \geq 1$$
$$= 0 \quad \text{otherwise}$$

$$T_2(a,b) = ab$$

$$T_{2.5}(a,b) = (a^{-1}+b^{-1}-1)^{-1}$$

$$T_3(a,b) = min(a,b)$$

Their corresponding DeMorgan dual T-conorms, denoted by $S_i(a,b)$, are defined as

$$S_i(a,b) = 1 - T_i(1-a, 1-b)$$

These five calculi provide the user with an ability to choose the desired uncertainty calculus starting from the most conservative ($T_1$) to the most liberal ($T_3$). $T_1$ ($T_3$) is the most conservative (liberal) T-norm in the sense that for the same input certainty ranges of facts and rule sufficiency and necessity measures, $T_1$ ($T_3$) shall yield the minimum (maximum) degree of confirmation of the conclusion. For each calculus (represented by the above five T-norms), the following four operations have been defined in RUM:

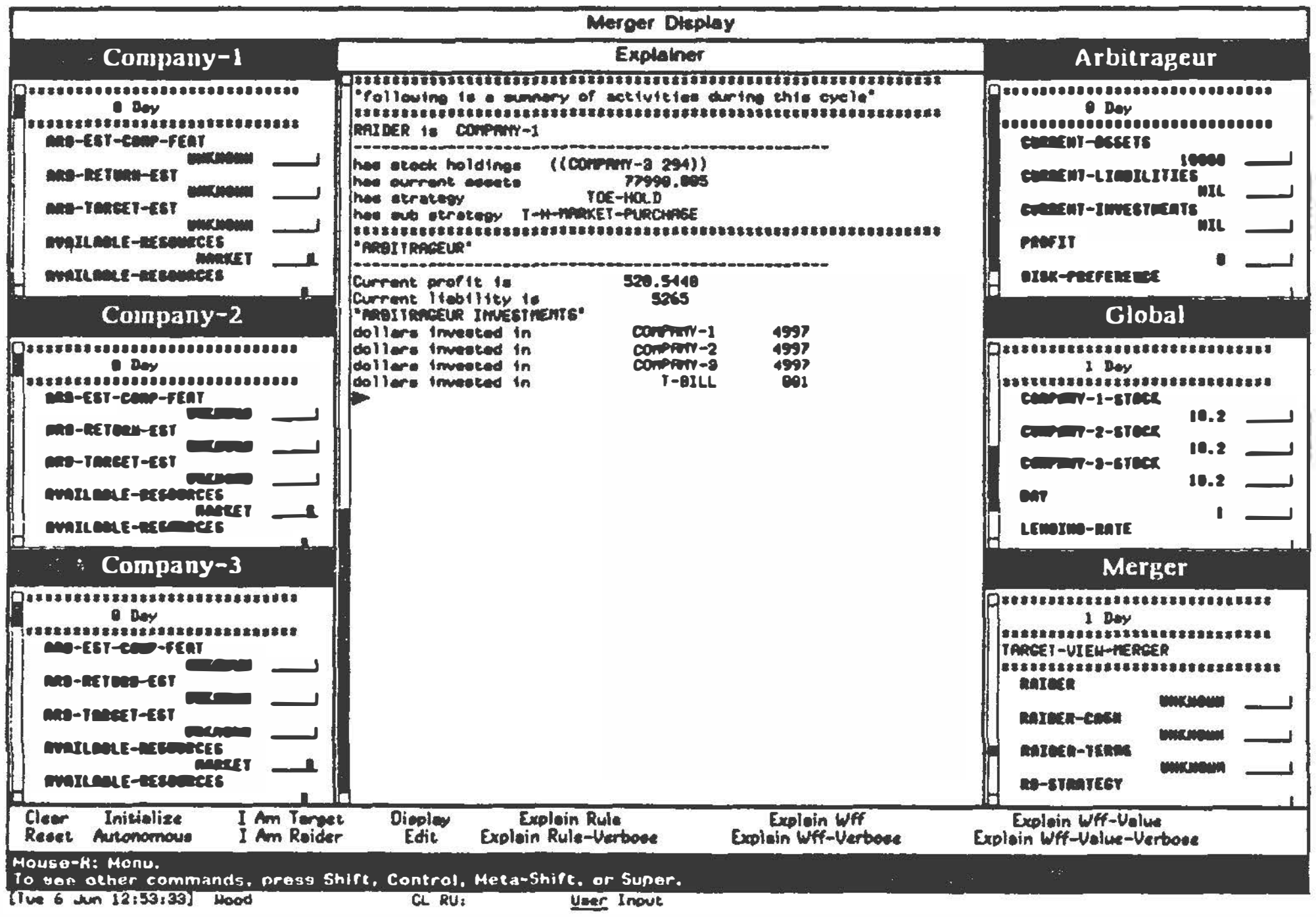

Figure 2: MARS User Interface

**Antecedent Evaluation.** To determine the aggregated certainty range [b, B] of the n clauses in the antecedent of a rule, when the certainty range of the ith clause is given by $[b_i, B_i]$:

$$[b, B] = [T_i(b_1, b_2, \ldots, b_n), T_i(B_1, B_2, \ldots, B_n)]$$

**Conclusion Detachment: Modus Ponens.** To determine the certainty range, [c,C] of the conclusion of a rule, given the aggregated certainty range, [b,B] of the rule premise and the rule sufficiency, $s$ and rule necessity, $n$:

$$[c, C] = [T_i(s, b), 1 - (T_i(n, (1 - B)))]$$

**Conclusion Aggregation.** To determine the consolidated certainty range [d, D], of a conclusion when it is supported by $m$ ($m > 1$) paths in the rule deduction graph, i.e., by $m$ rule instances, each with the same conclusion aggregation T-norm operator. If $[c_i, C_i]$ represents the certainty range of the same conclusion inferred by the ith proof path (rule instance), then

$$[d, D] = [S_i(c_1, c_2, \ldots c_m), 1 - S_i(C_1, C_2, \ldots, C_m)]$$

**Source Consensus.** To determine the certainty range, $[L_{tot}(A), U_{tot}(A)]$ of the same evidence, A, obtained by fusing the certainty ranges, $[L_i(A), U_i(A)]$, of the *ith* information source out of a total of $n$ different possible information sources:

$$[L_{tot}(A), U_{tot}(A)] = [Max_{i=1,\ldots,n} L_i(A), Min_{i=1,\ldots,n} U_i(A)]$$

The theory of RUM is anchored on the semantics of many-valued logics [Bon87b]. Unlike other probabilistic systems, RUM's reasoning mechanism is possibilistic. References [Bon87b], [BGD87] describe a comparison of RUM with other reasoning with uncertainty systems, such as Modified Bayesian [DHN76], Certainty Factors [SB75], [Hec86], Dempster-Shafer [Dem67], [Sha76], and Fuzzy logic [Zad65].

### 4.3 Integrating RBR and CBR in MARS

**RBR in MARS** MARS is implemented using RUM [BGD87] and KEE. RUM is implemented on KEE but uses its own rule system (only the KEE data structure and graphical interface facilities are used by RUM). These rules are "plausible rules" rather than strict implications and incorporate a sophisticated uncertainty calculus, as described in Section 4.2.

RUM offers both backward and forward processing. The expressiveness of RUM is enhanced by two other functionalities: the context mechanism and belief revision. The context represents the set of preconditions determining the rule's applicability to a given situation. This mechanism provides an efficient screening of the knowledge base by focusing the inference process on small rule subsets. The context of a rule forms an integral part of the RUM rule template. RUM's belief revision is essential to the dynamic aspect of the domain. The belief revision mechanism detects changes in the input, keeps track of the dependency of the intermediate and final conclusions on these inputs, and maintains the validity of these inferences. For any conclusions made by a rule, the mechanism monitors the changes in the certainty measures that constitute the conclusion's support. Rules in the MARS knowledge base are organized in hierarchical rule-classes. The reader is referred to [BGD87], [Bon87b] for more details on RUM and to [BD90] for details on the MARS rule base.

**CBR in MARS** We will now turn our attention to the CBR component in MARS. We will focus our discussion on four important issues related to CBR: representation & interpretation, need, matching & relevance, and integration.

- **Representation & Interpretation**: This issue is concerned with how the input problem/goal and the various cases in the case library are represented. Are they stored in an interpreted or uninterpreted (i.e., analyses of relevant features and event sequences are required) format? Is the representation complete or is it partial (i.e., some learning/non-monotonic reasoning is required)? What is the data/memory structure (rules/frames/MOPS [RS89a]) used?
- **Need**: This issue refers to determining whether the use of CBR is required for solving the input problem/goal. It follows from the observation that CBR is relevant only for certain kinds of problems/goals, e.g., in the domain of M&A, it is most relevant for the legal and tax aspects of the deal.
- **Matching & Relevance**: This issue is concerned with finding cases (from the case library) which are most similar to the input and determining which case is most relevant to solving/justifying the input problem/goal.
- **Integration**: How well can CBR be integrated into the overall problem solving structure? Until recently, this issue has been of little concern as most research in CBR has been done in isolation from other reasoning methodologies.

A review of the literature reveals that the issues of *representation & interpretation* and *matching & relevance* have received more discussion than the other two equally important issues of *need* and *integration*. We explain below how each of these issues has been addressed in MARS.

**Representation and Interpretation.** Given our intention to integrate RBR and CBR within the common architecture of rules, we have decided to represent individual cases in the MARS case library as RUM rules. The stored representation of cases consists of RUM rule templates. For example, the part of the Mobil-Marathon case elaborated upon in Section 3.2 would be represented in the following (pseudo English & Lisp) form:

```
                    (CASE 1)

IF   (similar-industry ?raider ?target) AND (Ti)
     (large-merged-national-market
                        ?raider ?target) AND (Ti)
     (significant-local-dominance ?raider ?target)
THEN significant chance (sufficiency)
     (anti-trust-success ?raider ?target)
```

where (`Ti`) is the particular T-norm operator chosen for conjunction of the three rule premises. Each premise (here) is a call to a procedure which returns an interval valued certainty measure (see Section 4.2) when the variables *?raider* and *?target* have been instantiated to a particular raider and target. The sufficiency measure, ***significant chance***, gives the degree to which the conjunction of the three premises is relevant for determining the success of the anti-trust suit in this case. (The necessity measure has been omitted for clarity). It should also be noted that Mobil and Marathon have been replaced by the role variables *?raider* and *?target* respectively.

The Mobil-Marathon case shall have many other such RUM rule templates to represent various aspects and events. A case library shall also have descriptions (rule templates) for other cases (e.g., the Brown Shoe and Kinney Co. case). In MARS a hierarchical structure is imposed on the case library containing various RUM rule templates. For example, the case library can have at the top level two divisions, one containing cases pertaining to *defensive* strategies and the other related to *attack* strategies. Within the defensive strategies category, we can have sub-categories for cases related to different types of defensive strategies (e.g., ***pac-man***, ***greenmail*** and ***anti-trust***). Going further down the sub- category for anti-trust defensive strategy cases, we can have sub-sub-categories for cases related to ***market dominance***, ***efficiency*** and ***foreign competition***. The example rule template, CASE 1, (described above) would be contained in the market dominance category.

Recall that a RUM rule (Section 4.3) has a context which keeps track of the environment in which that rule is activated. The context of rules used for RBR is used to efficiently index into the hierarchical structure imposed on the case library. For example, if the rule context indicates that anti-trust success is being evaluated, only related cases shall be retrieved. This shall become clearer below in the next few sub-sections. Rule templates representing cases also have contexts (not shown in CASE 1) and these are useful in determining the relevance of the cases to the input problem/goal.

To summarize, cases are stored in an interpreted, rule template format with a hierarchical, functional structure imposed on the case library. The uncertainty mechanism of RUM is utilized in two ways: first, to represent the relative importance of the premises for the conclusion (by the choice of Ti) and second, to represent the relevance of the premises to the conclusion (via the sufficiency and necessity measures). It is of course important to consider means to obtain the interpreted rule templates from available data. This is possible in the domain under consideration. Consider the ruling of Judge Manos in the Mobil- Marthon case which outlined detailed reasons for the judgement. An intelligent information retrieval system should be able to analyze such natural language data and extract the reasons (interpretations) for events and actions. SCISOR [JR90] is an intelligent natural language system which can perform this function. Reference [BD90] outlines this integration of SCISOR and MARS. In the absence of such recorded descriptions of events/actions and their interpretations, some inductive learning programs shall have to used to obtain the interpreted rule templates.

**Need for CBR.** It is important to recognize two points. First, CBR is important only for certain problems and goals. It is not useful to always consider CBR. For example, in the domain of M&A, CBR is useful primarily for structuring the legal and tax aspects of the deal. For some other aspects, such as use of statistical financial models, it makes little sense to include CBR. Second, CBR may be only one approach (proof path) to the solution of a goal/problem. There are (usually) other approaches (or proof paths) to the same goal/problem and it is important to consider the contributions of all paths, proportional to their relative importance. This aspect is significant, as most research in CBR has considered it in isolation till now. In MARS, the inference process can be considered as the traversal of paths in a rule graph. Premises, qualified by certainty intervals, combine (using RUM's uncertainty calculi) to generate conclusions (also qualified by certainty intervals) which either again act as premises for other rules or generate final conclusions. A simple rule graph is shown in Figure 3. Since CBR is just one path for proving a certain goal, a rule to this effect is included in the rule structure whenever the expert feels that CBR is important for the present goal. For example, consider a hypothetical M&A deal, M1, being analyzed by MARS. In the evaluation of the possibility of success of an anti-trust suit in M1, a rule would be added:

```
          (RULE 1)

IF similar anti-trust precedent  exists
THEN high chance (sufficiency)
  anti-trust successful in M1
```

It should be noted that there shall be other rules (proof paths) also that either confirm/disconfirm the current goal under consideration. For example, there might be a rule:

```
          (RULE 2)

IF target has strong political lobby
THEN it is likely (sufficiency)
  anti-trust successful in M1
```

The above rules represent two different proof paths, each contributing to the determination of the goal *"anti-trust successful in M1"* (see Figure 3). The ***conclusion aggregation*** and ***source consensus*** operations (see 3.2.2) determine the relative contributions of RULE-1 and RULE-2 to the final conclusion of *"anti-trust successful in M1"*.

**Matching and Relevance.** The matching and relevance process is operationalized by instantiating the case rule templates in the case library to the situation of the current world, M1. This process converts the rule templates to RUM rules which can be used in the reasoning process of M1 and at the same time determines the degree of relevance of the previous cases to M1. Thus if case 1 (rule template) were instantiated to M1 world conditions, the variables *?raider* and *?target* would be instantiated to the raider and target respectively in M1, and each of the three premises shall be evaluated to yield certainty ranges which give the degree to which the premises of the case are true in the current M1 world. If they are not relevant (true), a very low confirmation for the premises shall be obtained and vice versa. Using the uncertainty calculi of RUM, case 1 shall yield a conclusion with a certainty range which is the degree to which that case is relevant to M1. As there shall be many cases for the same conclusion (e.g., successful anti-trust cases) in the case library, an aggregated value of the relevance of all the previous cases can be obtained using the conclusion aggregation and source consensus operations of RUM's uncertainty calculi. The node labelled ***" anti-trust success"*** represents the aggregated contribution of various cases for determining the success of an anti-trust suit in M1.

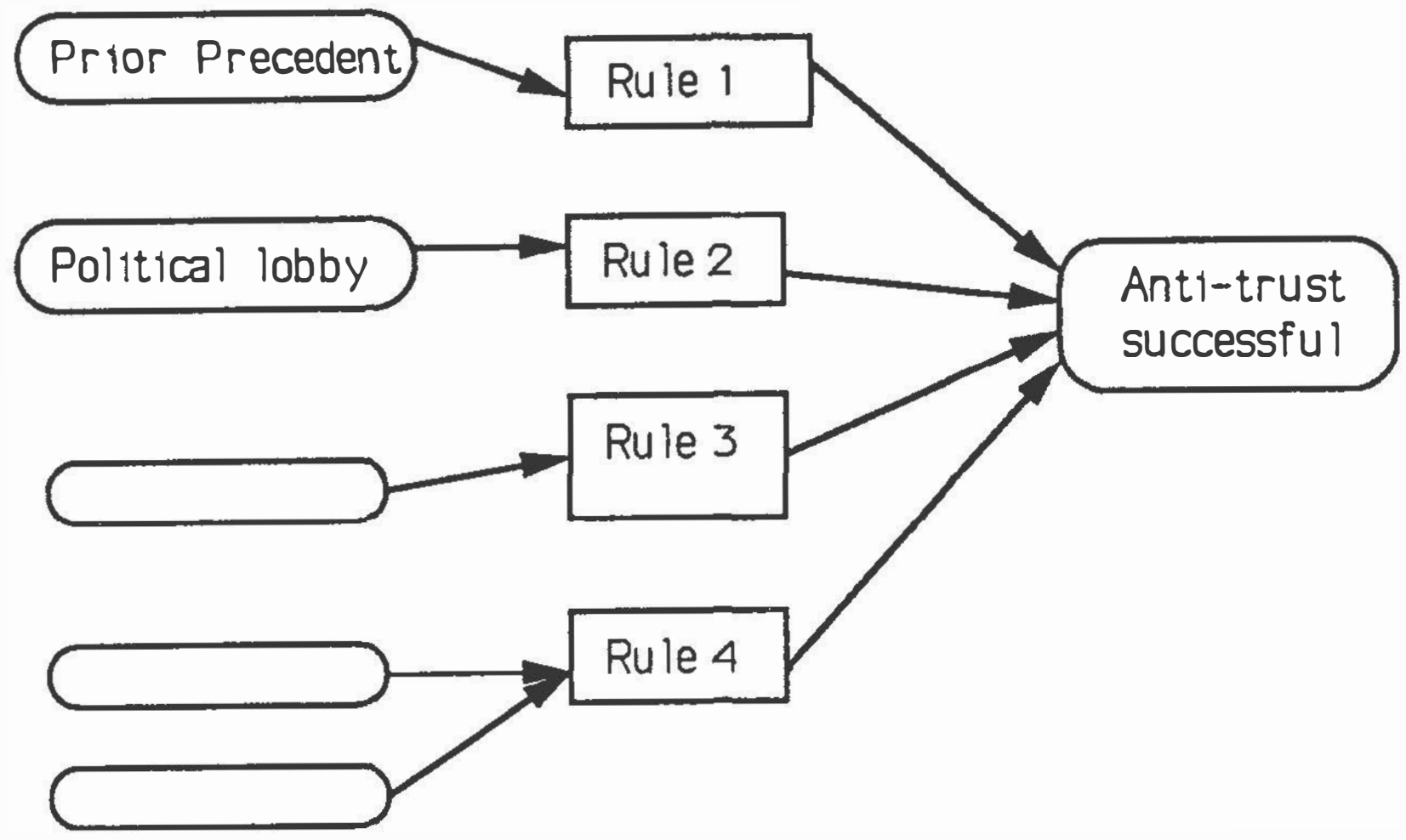

Figure 3: Example of a Simple Rule Graph

This process of matching can be alternatively understood by noting that the *necessity measure* $N(p \mid d)$ represents the degree of semantic entailment of a pattern descriptor p given a datum d. The *possibility measure* $P(p \mid d)$ represents the degree of intersection between the same pattern and datum. Thus, the interval defined by $[N(p \mid d), P(p \mid d)]$ represents the lower and upper bounds on the degree of matching between such pattern and data. This interval is the same as the interval valued certainty ranges obtained when premises of case rule templates are instantiated to the current world conditions.

**Integration.** The above sub-sections have outlined the details of the integration of RBR and CBR in MARS. The last detail in the integration process is observing that the node labelled *"Prior Precedent"* in Figure 3 is automatically expanded by the backward chainer of the RBR when evaluating the truth value of the node *"Anti-trust Success"* (Figure 4). This expansion is shown in Figure 4.

Thus to summarize the process briefly:

- Cases are stored as rule templates (CASE 1).
- If CBR is important, a rule to this effect (RULE 1) is added.
- Case rule templates are instantiated automatically while evaluating the premise of rules like (RULE 1). Rule contexts are used for indexing into the hierarchical structure of the case library.
- Instantiation of case rule templates (CASE 1) automatically determines relevance and matching using the T-norm based uncertainty calculi of RUM.

Finally, we would like to emphasize the seamless and compact integration of RBR and CBR. No changes need to be made in the inference engine of either RUM or MARS (which remains the same whether CBR is used or not).

# 5 Conclusion

## 5.1 Comparisons with Related Research

As mentioned earlier, there has been very little research in combining RBR and CBR. The only work in this area that the authors are aware of is that of Rissland and Skalak [RS89b]. Their approach however, is very different. While working in the legal domain of *statutory interpretation*, they have built a system called CABARET, whose architecture consists of two co-equal reasoners, one a RBR and the other a CBR, with a separate agenda based controller. The central controller contains heuristics to direct and interleave the two modes of reasoning and to post and prioritize tasks for each controller. This effort should be credited for being the first in addressing such an important issue. However, we feel that it is difficult to choose the right heuristics for the controller and to design it to perform correctly and adequately in different, complex domains. Our approach to integrating RBR and CBR also provides a treatment of uncertainty and approximate matching between input and cases, which is not available in CABARET.

**Contributions and Limitations** We feel that the primary contribution of this paper has been to illustrate the compact and seamless integration of RBR and CBR as implemented in MARS. Both RBR and CBR are very important reasoning methodologies and it is surprising that there has been such little prior work in integrating the two. We hope that this paper represents a major effort in that direction. Both RBR and CBR are required for solving complex real world problems. By choosing RBR as the base architecture for integration, we have illustrated a method for adding more power to rule based systems, i.e., expanding their inference capabilities. Our architecture treats the contributions of CBR and RBR simultaneously and proportionately (according to their relative importance) as separate proof paths to a conclusion. This does not require the use of any special heuristics or agendas. As shown, no changes have to be made to the inference engine of RBR to accommodate CBR. Furthermore, this seamless integration is anchored on common possibilistic semantics for both CBR and RBR. The methodology presented in this paper is general and also applies to RBR without uncertainty (where rules and facts are special cases of RUM rules and facts) and to both problem solving CBR and precedent based CBR (as long as the base architecture is rule based).

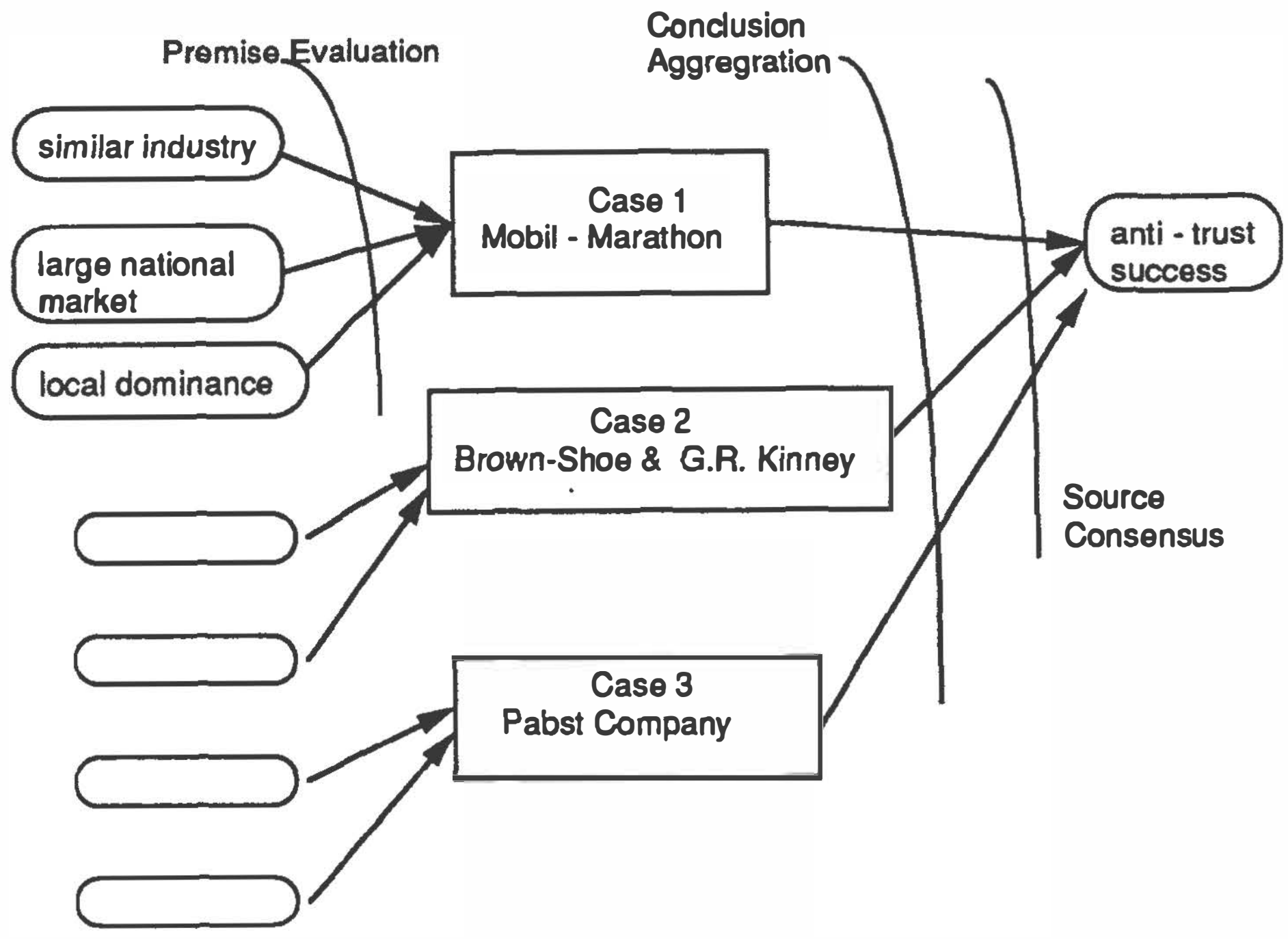

Figure 4: Expansion of Rule Graph

We will conclude our discussion by noting some of the limitations of the methodology described in this paper and by proposing future efforts aimed at strengthening this approach. The case library consists of interpreted rule templates (cases). The process of interpretation of data to obtain such rule templates, though possible, is non-trivial. We are looking into the use of SCISOR [JR90] for such purposes. Also, the case library at present has to be necessarily incomplete (as it is not possible to represent all possible reasons for all possible events/actions). Currently, we have chosen to represent only important events/actions in a case. We are now investigating the use of machine learning and non-monotonic reasoning techniques for handling this incompleteness. A possibility in this connection, is to have a system such as SCISOR "lazily" interpreting case data on demand. This approach, however requires a more thorough treatment of issues related to dynamic analysis of cases, efficiency, etc. All these issues will be the focus of our research goals for the forthcoming future.